\documentclass[runningheads]{llncs}
\usepackage{graphicx}
\usepackage{hyperref}

\usepackage{amsmath}
\usepackage{mathrsfs}
\usepackage{lmodern}
\usepackage{amssymb}
\usepackage{tkz-graph}          
\usepackage{caption}
\usepackage{subfig}
\usepackage{csquotes}
\usepackage{graphicx}
\usepackage{multirow}

\DeclareMathOperator*{\argmin}{arg\,min}
\begin{document}

\title{Part-based approximations for morphological operators using asymmetric auto-encoders}
\titlerunning{Approximations for morphological operators using asymmetric auto-encoders}
%
\author{Bastien Ponchon\inst{1,2} \and
Santiago Velasco-Forero\inst{1} \and
Samy Blusseau\inst{1} \and
Jes\'us Angulo\inst{1} \and
Isabelle Bloch\inst{2}}
\authorrunning{B. Ponchon et al.}
%
\institute{Centre for Mathematical Morphology, Mines ParisTech, PSL Research University, France \and
LTCI, T\'el\'ecom ParisTech, Universit\'e Paris-Saclay, France}
\maketitle              
\begin{abstract}
  This paper addresses the issue of building a part-based
  representation of a dataset of images. More precisely, we look for a
  non-negative, sparse decomposition of the images on a reduced set of
  atoms, in order to unveil a morphological and interpretable
  structure of the data. Additionally, we want this decomposition to
  be computed online for any new sample that is not part of the
  initial dataset. Therefore, our solution relies on a sparse,
  non-negative auto-encoder where the encoder is deep (for accuracy)
  and the decoder shallow (for interpretability). This method compares
  favorably to the state-of-the-art online methods on two datasets
  (MNIST and Fashion MNIST), according to classical metrics and to a
  new one we introduce, based on the invariance of the representation
  to morphological dilation.

\end{abstract}
\section{Introduction}

Mathematical morphology is strongly related to the problem of data
representation. Applying a morphological filter can be seen as a test
on how well the analyzed element is represented by the set of
invariants of the filter. For example, applying an opening by a
structuring element $B$ tells how well a shape can be represented by
the supremum of translations of $B$. The morphological skeleton
\cite{maragos1986morphological,soille2013morphological} is a typical
example of description of shapes by a family of building blocks,
classically homothetic spheres. It provides a disjunctive
decomposition where components~- for example, the spheres - can only
contribute positively as they are combined by supremum. A natural
question is the optimality of this additive decomposition according to
a given criterion, for example its sparsity~- the number of components
needed to represent an object. Finding a sparse disjunctive (or
part-based) representation has at least two important features: first,
it allows \emph{saving resources} such as memory and computation time
in the processing of the represented object; secondly, it provides a
\emph{better understanding} of this object, as it reveals its most
elementary components, hence operating a dimensionality reduction that
can alleviate the issue of model over-fitting. Such representations
are also believed to be the ones at stake in human object recognition
\cite{Tanaka2003ColumnsFC}.

Similarly, the question of finding a sparse disjunctive representation
of a whole database is also of great interest and will be the main
focus of the present paper. More precisely, we will approximate such a
representation by a non-negative, sparse linear combination of
non-negative components, and we will call \emph{additive} this
representation. Given a large set of images, our concern is then to
find a smaller set of non-negative image components, called
dictionary, such that any image of the database can be expressed as an
additive combination of the dictionary components. As we will review
in the next section, this question lies at the crossroad of two
broader topics known as sparse coding and dictionary learning
\cite{DBLP:journals/corr/MairalBP14}.

Besides a better understanding of the data structure, our approach is
also more specifically linked to mathematical morphology
applications. Inspired by recent work
\cite{angulo:hal-00658963,Angulo_Velasco_2017}, we look for image
representations that can be used to efficiently calculate
approximations to morphological operators. The main goal is to be able
to apply morphological operators to massive sets of images by applying
them only to the reduced set of dictionary images. This is especially
relevant in the analysis of remote sensing hyperspectral images where
different kinds of morphological decomposition, such as morphological
profiles \cite{Morphological_profiles} are widely used. For reasons
that will be explained later, sparsity and non-negativity are sound
requirements to achieve this goal. What is more, whereas the
representation process can be learned offline on a training dataset,
we need to compute the decomposition of any new sample
\emph{online}. Hence, we take advantage of the recent advances in
deep, sparse and non-negative auto-encoders to design a new framework
able to learn part-based representations of an image database,
compatible with morphological processing.

The existing work on non-negative sparse representations of images are
reviewed in Section 2, that stands as a baseline and motivation of the
present study. Then we present in Section 3 our method before showing
results on two image datasets
(MNIST~\cite{lecun-mnisthandwrittendigit-2010} and Fashion
MNIST~\cite{fashionMNIST}) in Section 4, and show how it compares to
other deep part-based representations. We finally draw conclusions and
suggest several tracks for future work in Section 5. The code for
reproducing our experiments is available online\footnote{For code
  release, visit
  \url{https://gitlab.telecom-paristech.fr/images-public/asymae_morpho}}.

\section{Related work}
\label{sec:related-work}

\subsection{Non-negative sparse mathematical morphology}
\label{sec:non-negative-sparse-mm}

The present work finds its original motivation
in~\cite{Angulo_Velasco_2017}, where the authors set the problem of
learning a representation of a large image dataset to quickly compute
approximations of morphological operators on the images. They find a
good representation in the sparse variant of Non-negative Matrix
Factorization (sparse NMF)~\cite{Hoyer2004}, that we present
hereafter.

Consider a family of $M$ images (binary or gray-scale)
$\mathbf{x}^{(1)}$, $\mathbf{x}^{(2)}$, ..., $\mathbf{x}^{(M)}$ of $N$
pixels each, aggregated into a $M \times N$ data matrix
$\mathbf{X}=(\mathbf{x}^{(1)}, \mathbf{x}^{(2)}, ...,
\mathbf{x}^{(M)})^T$ (the $i^{th}$ row of $\mathbf{X}$ is the
transpose of $\mathbf{x}^{(i)}$ seen as a vector). Given a feature
dimension $k\in\mathbb{N}^*$ and two numbers $s_H$ and $s_W$ in
$[0,1]$, a sparse NMF of $\mathbf{X}$ with dimension $k$, as defined
in~\cite{Hoyer2004}, is any solution of the
problem 
\begin{equation}
    \label{eq:sNMF}
    \begin{array}{ccc}
      \mathbf{H}\mathbf{W} = \argmin ||\mathbf{X} - \mathbf{H}\mathbf{W}||_2^2
      &   \;\; \text{s.t.}\;
      & \left\lbrace\begin{array}{c}
                      \mathbf{H}\in\mathbb{R}^{M \times k}, \mathbf{W} \in \mathbb{R}^{k \times N}\\
                      \mathbf{H}\geq 0, \; \mathbf{W}\geq 0 \\
                      \sigma(\mathbf{H}_{:,j})=s_H, \sigma(\mathbf{W}_{j,:})=s_W, \; 1\leq j\leq k
                    \end{array}\right.
    \end{array}
  \end{equation}
  where the second constraint means that both $\mathbf{H}$ and
  $\mathbf{W}$ have non-negative coefficients, and the third
  constraint imposes the degree of sparsity of the columns of
  $\mathbf{H}$ and lines of $\mathbf{W}$ respectively, with $\sigma$
  the function defined by
\begin{equation}
\label{eq:hoyer-sparsity}
\forall \mathbf{v}\in\mathbb{R}^p,\;\;\;    \sigma(\mathbf{v}) = \frac{\sqrt{p}-||\mathbf{v}||_1/||\mathbf{v}||_2}{\sqrt{p}-1}.
\end{equation}
Note that $\sigma$ takes values in $[0,1]$. The value
$\sigma(\mathbf{v}) = 1$ characterizes vectors $\mathbf{v}$ having a
unique non-zero coefficient, therefore the sparsest ones, and
$\sigma(\mathbf{v}) = 0$ the vectors whose coefficients all have the
same absolute value. Hoyer~\cite{Hoyer2004} designed an algorithm to
find at least a local minimizer for the problem (\ref{eq:sNMF}), and
it was shown that under fairly general conditions (and provided the
$L_2$ norms of $\mathbf{H}$ and $\mathbf{W}$ are fixed) the solution
is unique~\cite{theis2005first}.

In the terminology of representation learning, each row
$\mathbf{h}^{(i)}$ of $\mathbf{H}$ contains the \emph{encoding} or
\emph{latent features} of the input image $\mathbf{x}^{(i)}$, and
$\mathbf{W}$ holds in its rows a set of $k$ images called the
dictionary. In the following, we will use the term \textit{atom
  images} or \textit{atoms} to refer to the images
$\mathbf{w}_j = \mathbf{W}_{j,:}$ of the dictionary. As stated by
Equation~\eqref{eq:sNMF}, the atoms are combined to approximate each
image $\mathbf{x^{(i)}} := \mathbf{X}_{i,:}$ of the dataset. This
combination also writes as follows:
\begin{equation}
\label{eq:linear-approximation}
\begin{array}{lrcl}
  \forall i \in \{1,...,M\},\;\;\; &  
                                     \mathbf{x}^{(i)} \approx \mathbf{\hat{x}}^{(i)}  & = & \mathbf{H}_{i,:}\mathbf{W} = \mathbf{h}^{(i)}\mathbf{W} =  \sum_{j=1}^k h_{i,j}\mathbf{w}_j. 
\end{array}
\end{equation}
The assumption behind this decomposition is that the more similar the
images of the set, the smaller the required dimension to accurately
approximate it. Note that only $k(N + M)$ values need to be stored or
handled when using the previous approximation to represent the data,
against the $NM$ values composing the original data.

By choosing the sparse NMF representation, the authors
of~\cite{Angulo_Velasco_2017} aim at approximating a morphological
operator $\phi$ on the data $\mathbf{X}$ by applying it to the atom
images $\mathbf{W}$ only, before projecting back into the input image
space. That is, they want
$\phi(\mathbf{x}^{(i)})\approx \Phi(\mathbf{x}^{(i)})$, with
$\Phi(\mathbf{x}^{(i)})$ defined by
\begin{equation}
    \label{eq:max-approx}
    \Phi(\mathbf{x}^{(i)}) := \sum_{j=1}^k h_{i,j}\phi(\mathbf{w}_j).
\end{equation}
The operator $\Phi$ in Equation~\eqref{eq:max-approx} is called a
\textbf{part-based approximation} to $\phi$.  To understand why
non-negativity and sparsity allow hoping for this approximation to be
a good one, we can point out a few key arguments. First, sparsity
favors the support of the atom images to have little pairwise
overlap. Secondly, a sum of images with disjoint supports is equal to
their (pixel-wise) supremum. Finally, dilations commute with the
supremum and, under certain conditions that are favored by sparsity it
also holds for the erosions. To precise this, let us consider a flat,
extensive dilation $\delta_B$ and its adjoint anti-extensive erosion
$\varepsilon_B$, $B$ being a flat structuring element. Assume
furthermore that for any $i \in [1,M], (j,l)\in [1,k]^2$ with
$j\neq l$,
$\delta_B(h_{i,j}\mathbf{w}_j) \bigwedge
\delta_B(h_{i,l}\mathbf{w}_l)=0$. Then on the dataset $\mathbf{X}$,
$\delta_B$ and $\varepsilon_B$ are equal to their approximations as
defined by Equation~\eqref{eq:max-approx}, that is to say:
$$
\begin{array}{ccl}
  \delta_B(\mathbf{x}^{(i)}) & = & \delta_{B}\left(\sum_{j=1}^k h_{i,j}\mathbf{w}_j\right)
                                   = \delta_{B}\left(\underset{j \in [1,k]}{\bigvee}h_{i,j}\mathbf{w}_j\right)
                                   = \underset{[1,k]}{\bigvee}\delta_{B}(h_{i,j}\mathbf{w}_j)\\
                             & = & \sum_{j=1}^k \delta_{B}(h_{i,j}\mathbf{w}_j) = \sum_{j=1}^k h_{i,j}\delta_{B}(\mathbf{w}_j) := D_B(\mathbf{x}^{(i)})
\end{array}
$$
and similarly, since
$\delta_B(x)\wedge\delta_B(y) = 0 \Rightarrow \varepsilon_B(x\vee y)=
\varepsilon_B(x)\vee\varepsilon_B(y)$ for $\delta_B$ extensive, we
also get
$\varepsilon_B(\mathbf{x}^{(i)}) = \sum_{j=1}^k
h_{i,j}\varepsilon_{B}\left(\mathbf{w}_j\right) :=
E_B(\mathbf{x}^{(i)}).$ It follows that the same holds for the opening
$\delta_B\varepsilon_B$. The assumption we just made is obviously too
strong and unlikely to be verified, but this example helps realize
that the sparser the non-negative decomposition, the more disjoint the
supports of the atom images and the better the approximation of a flat
morphological operator.

\begin{figure}[htbp]
\begin{center}
\includegraphics[width=0.8\linewidth]{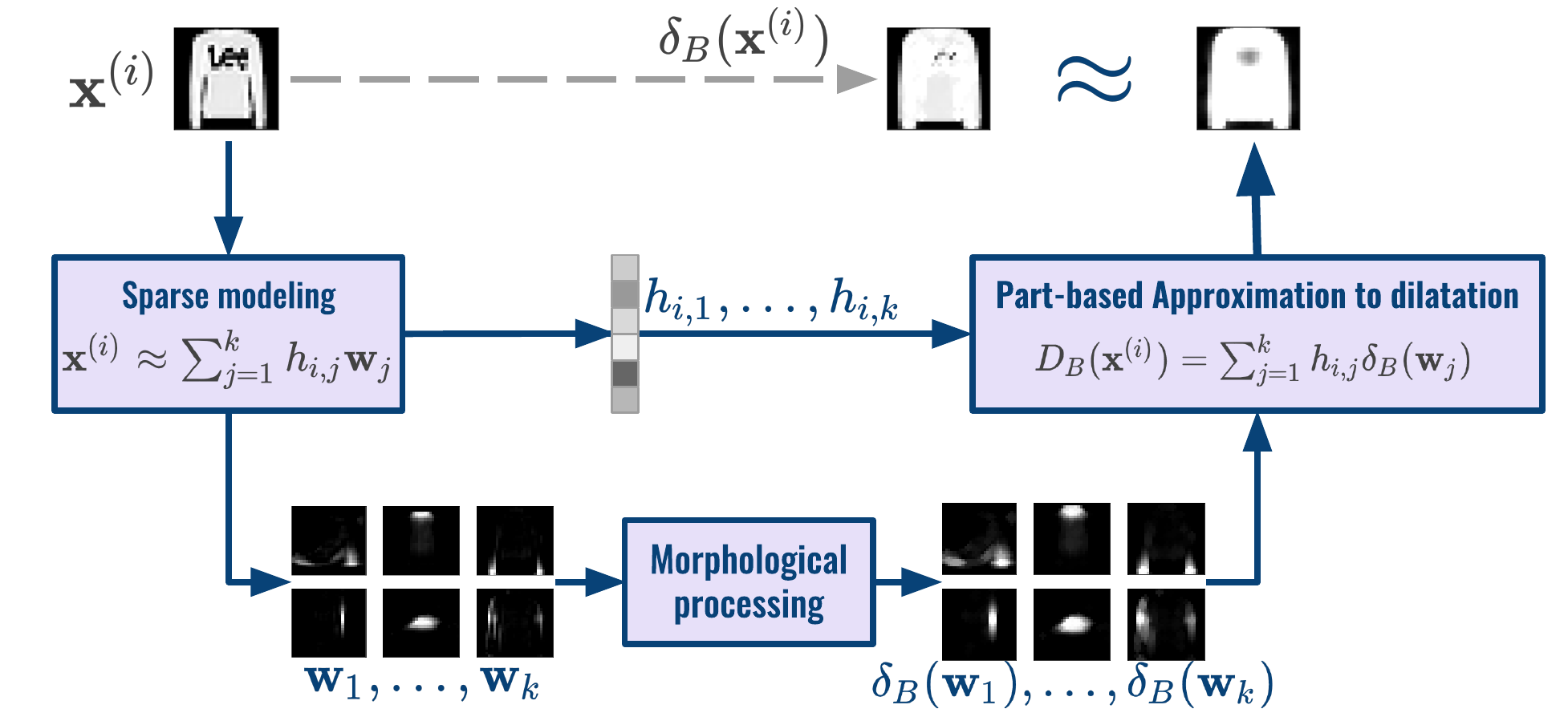}\end{center}
   \caption{Process for computing the part-based-approximation of dilation.}
\label{fig:max_approx_process}
\end{figure}

As a particular case, in this paper we will focus on part-based
approximations of the dilation by a structuring element $B$, expressed
as:
\begin{equation}
  D_{B}(\mathbf{x}^{(i)}) := \sum_{j=1}^k h_{i,j}\delta_{B}(\mathbf{w}_j),
\end{equation}
that we will compare with the actual dilation of our input images to
evaluate our model, as shown in Figure~\ref{fig:max_approx_process}.

\subsection{Deep auto-encoders approaches}
\label{sec:autoencoder-approaches}
The main drawback of the NMF algorithm is that it is an
\textit{offline} process, the encoding of any new sample with regards
to the previously learned basis $\mathbf{W}$ requires either to solve
a computationally extensive constrained optimization problem, or to
release the Non-Negativity constraint by using the pseudo-inverse
$\mathbf{W^+}$ of the basis. The various approaches proposed to
overcome this shortcoming rely on Deep Learning, and especially on
deep auto-encoders, which are widely used in the representation
learning field, and offer an \textit{online} representation process.

\begin{figure}[htbp]
\begin{center}
  \includegraphics[width=0.7\linewidth]{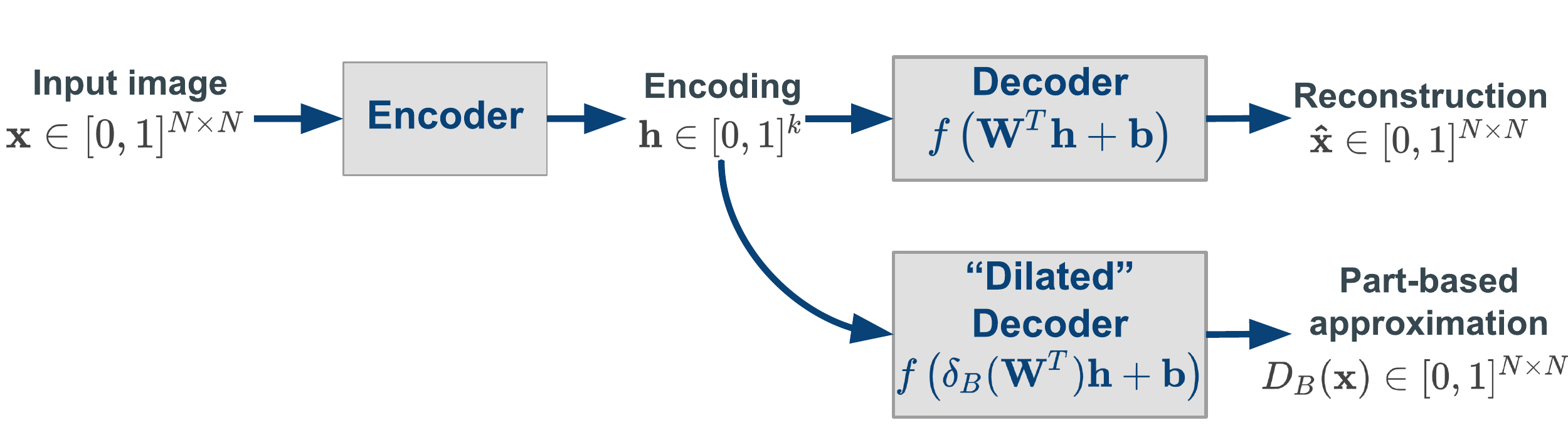}\end{center}
\caption{The auto-encoding process and the definition of part-based
  approximation to dilation by a structuring element $B$ in this
  framework.}
\label{fig:AE_framework}
\end{figure}

An auto-encoder, as represented in Figure~\ref{fig:AE_framework}, is a
model composed of two stacked neural networks, an encoder and a
decoder whose parameters are trained by minimizing a loss function. A
common example of loss function is the mean square error (MSE) between
the input images $\mathbf{x}^{(i)}$ and their reconstructions by the
decoder $\mathbf{\hat{x}}^{(i)}$:
\begin{equation}
    L_{AE} = \frac{1}{M} \sum_{i=1}^M L(\mathbf{x}^{(i)}, \mathbf{\hat{x}}^{(i)}) 
    = \frac{1}{M} \sum_{i=1}^M \frac{1}{N}|| \mathbf{\hat{x}}^{(i)} -\mathbf{x}^{(i)}||_2^2.
\end{equation}    
In this framework, and when the decoder is composed of a single linear
layer (possibly followed by a non-linear activation), the model
approximates the input images as:
\begin{eqnarray}
\label{eq:decoder}
    \mathbf{\hat{x}}^{(i)} &= f \left(\mathbf{b} + \mathbf{h}^{(i)}\mathbf{W}\right)
     &= f \left(\mathbf{b} + \sum_{j=1}^k h_{i,j}\mathbf{w}_j \right) 
\end{eqnarray}
where $\mathbf{h}^{(i)}$ is the encoding of the input image by the
encoder network, $\mathbf{b}$ and $\mathbf{W}$ respectively the bias
and weights of the linear layer of the decoder, and $f$ the (possibly
non-linear) activation function, that is applied pixel-wise to the
output of the linear layer. The output $\mathbf{\hat{x}}^{(i)}$ is
called the \textit{reconstruction} of the input image
$\mathbf{x}^{(i)}$ by the auto-encoder. It can be considered as a
linear combination of atom images, up to the addition of an offset
image $\mathbf{b}$ and to the application of the activation function
$f$. The images of our learned dictionary are hence the columns of the
weight matrix $\mathbf{W}$ of the decoder. We can extend the
definition of part-based approximation, described in
Section~\ref{sec:non-negative-sparse-mm}, to our deep-learning
architectures, by applying the morphological operator to these atoms
$\mathbf{w}_1$, ..., $\mathbf{w}_k$, as pictured by the
\textit{``dilated decoder"} in Figure~\ref{fig:AE_framework}. Note
that a central question lies in how to set the size $k$ of the latent
space. This question is beyond the scope of this study and the value
of $k$ will be arbitrarily fixed (we take $k=100$) in the following.

The NNSAE architecture, from Lemme \textit{et al.}
\cite{Lemme2012OnlineLA}, proposes a very simple and shallow
architecture for online part-based representation using linear encoder
and decoder with tied weights (the weight matrix of the decoder is the
transpose of the weight matrix of the encoder). Both the NCAE
architectures, from Hosseini-Asl \textit{et al.} \cite{7310882} and
the work from Ayinde \textit{et al.}
\cite{DBLP:journals/corr/abs-1802-00003} that aims at extending it,
drop this transpose relationship between the weights of the encoder
and of the decoder, increasing the capacity of the model. Those three
networks enforce the non-negativity of the elements of the
representation, as well as the sparsity of the image encodings using
various techniques.

\subsubsection{Enforcing sparsity of the encoding}\label{sec:sparsity}
\vspace{-.5cm} The most prevalent idea to enforce sparsity of the
encoding in a neural network can be traced back to the work of H. Lee
\textit{et al.} \cite{NIPS2007_3313}. This variant penalizes, through
the loss function, a deviation $S$ of the expected activation of each
hidden unit (\textit{i.e.} the output units of the encoder) from a low
fixed level $p$. Intuitively, this should ensure that each of the
units of the encoding is activated only for a limited number of
images. The resulting loss function of the sparse auto-encoder is
then:
\begin{equation}\label{eq:adding_sparse_regularization_term}
  L_{AE} = \frac{1}{M} \sum_{i=1}^M L(\mathbf{x}^{(i)}, \mathbf{\hat{x}}^{(i)} )  + \beta \sum_{j=1}^k S(p, \sum_{i=1}^{M}h_j^{(i)}),
\end{equation}
where the parameter $p$ sets the expected activation objective of each
of the hidden neurons, and the parameter $\beta$ controls the strength
of the regularization. The function $S$ can be of various forms, which
were empirically surveyed in~\cite{7280364}. The approach adopted by
the NCAE~\cite{7310882} and its
extension~\cite{DBLP:journals/corr/abs-1802-00003} rely on a penalty
function based on the KL-divergence between two Bernoulli
distributions, whose parameters are the expected activation and $p$
respectively, as used in~\cite{7310882}:
\begin{equation}\label{eq:KL_div_penalty_function}
  S(p, t_j) = KL(p,t_j) = p\log\frac{p}{t_j} + (1-p)\log\frac{1-p}{1-t_j} \; \; \mbox{ with }
  t_j =\sum_{i=1}^{M}h_j^{(i)}
\end{equation}
The NNSAE architecture~\cite{Lemme2012OnlineLA} introduces a slightly
different way of enforcing the sparsity of the encoding, based on a
parametric logistic activation function at the output of the encoder,
whose parameters are trained along with the other parameters of the
network.

\subsubsection{Enforcing non-negativity of the decoder weights}
\vspace{-0.5cm} For the NMF (Section~\ref{sec:non-negative-sparse-mm})
and for the decoder, non-negativity results in a part-based
representation of the input images. In the case of neural networks,
enforcing the non-negativity of the weights of a layer eliminates
cancellations of input signals. In all the aforementioned works, the
encoding is non-negative since the activation function at the output
of the encoder is a sigmoid. In the literature, various approaches
have been designed to enforce weight positivity. A popular approach is
to use an asymmetric weight decay, added to the loss function of the
network, to enact more decay on the negative weights that on the
positive ones. However this approach, used in both the
NNSAE~\cite{Lemme2012OnlineLA} and NCAE~\cite{7310882} architectures,
does not ensure that all weights will be non-negative. This issue
motivated the variant of the NCAE
architecture~\cite{DBLP:journals/corr/abs-1802-00003,Lemme2012OnlineLA},
which uses either the $L_1$ rather than the $L_2$ norm, or a smoothed
version of the decay using both the $L_1$ and the $L_2$ norms. The
source code of that method being unavailable, we did not use this more
recent version as a baseline for our study.

\section{Proposed model}

\begin{figure}[htbp]
\begin{center}
  \includegraphics[width=0.65\linewidth]{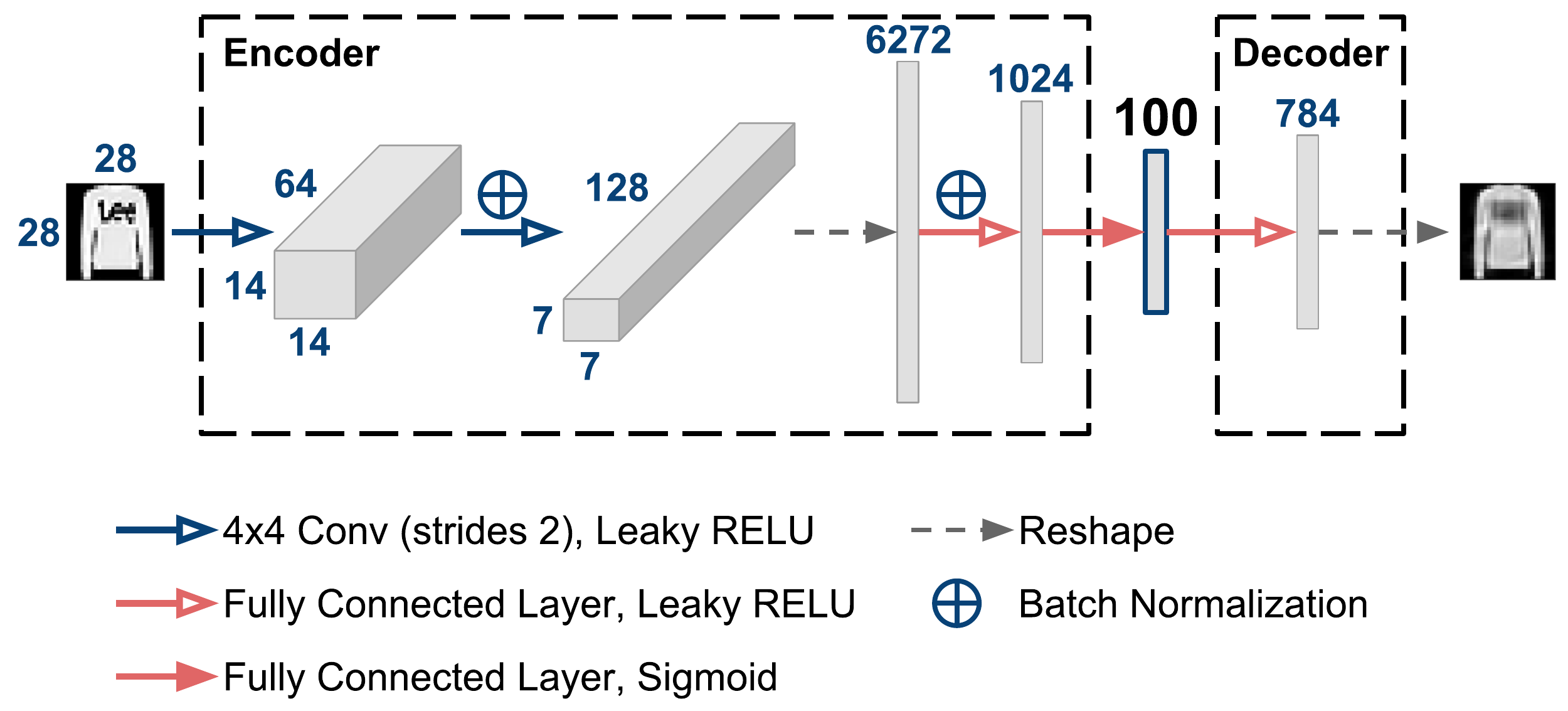}\end{center}
\caption{Our proposed auto-encoder architecture.}
\label{fig:asymAEinfoGAN_architecture}
\end{figure}

We propose an online part-based representation learning model, using
an asymmetric auto-encoder with sparsity and non-negativity
constraints.
As pictured in Figure~\ref{fig:asymAEinfoGAN_architecture}, our
architecture is composed of two networks: a deep encoder and a shallow
decoder (hence the asymmetry and the name of AsymAE we chose for our
architecture). The encoder network is based on the discriminator of
the infoGAN architecture introduced in \cite{infoGAN}, which was
chosen for its average depth, its use of widely adopted deep learning
components such as batch-normalization~\cite{batchnorm},
2D-convolutional layers \cite{DBLP:journals/corr/DosovitskiySB14} and
leaky-RELU activation function \cite{Maas2013RectifierNI}. It has been
designed specifically to perform interpretable representation learning
on datasets such as MNIST and Fashion-MNIST. The network can be
adapted to fit to larger images. The decoder network is similar to the
one presented in Figure~\ref{fig:AE_framework}. A Leaky-ReLU
activation has been chosen after the linear layer. Its behavior is the
same as the identity for positive entries, while it multiplies the
negative ones by a fixed coefficient $\alpha_{lReLU}=0.1$. This
activation function has shown better performances in similar
architectures~\cite{Maas2013RectifierNI}.  The sparsity of the
encoding is achieved using the same approach as in
\cite{DBLP:journals/corr/abs-1802-00003,7310882} that consists in
adding to the previous loss function the regularization term described
in Equations \eqref{eq:adding_sparse_regularization_term} and
\eqref{eq:KL_div_penalty_function}.

We only enforced the non-negativity of the weights of the decoder, as
they define the dictionary of images of our learned representation and
as enforcing the non-negativity of the encoder weights would bring
nothing but more constraints to the network and lower its capacity. We
enforced this non-negativity constraint explicitly by projecting our
weights on the nearest points of the positive orthant after each
update of the optimization algorithm (such as the stochastic gradient
descent). The main asset of this other method that does not use any
additional penalty functions, and which is quite similar to the way
the NMF enforces non-negativity, is that it ensures positivity of all
weights without the cumbersome search for good values of the
parameters the various regularization terms in the loss function.

\section{Experiments}\label{sec:results}
To demonstrate the goodness and drawbacks of our method, we have
conducted experiments on two well-known datasets
MNIST~\cite{lecun-mnisthandwrittendigit-2010} and Fashion
MNIST~\cite{fashionMNIST}. These two datasets share common features,
such as the size of the images ($28 \times 28$), the number of classes
represented ($10$), and the total number of images ($70000$), divided
in a training set of $60000$ images and a test set of $10000$
images. We compared our method to three baselines: the
sparse-NMF~\cite{Hoyer2004}, the NNSAE~\cite{Lemme2012OnlineLA}, the
NCAE~\cite{7310882}.  The three deep-learning models (AsymAE (ours),
NNSAE and NCAE) were trained until convergence on the training set,
and evaluated on the test set. The sparse-NMF algorithm was ran and
evaluated on the test set. Note that all models but the NCAE may
produce reconstructions that do not fully belong to the interval
$[0,1]$. In order to compare the reconstructions and the part-based
approximation produced by the various algorithms, their outputs will
be clipped between 0 and 1. There is no need to apply this operation
to the output of NCAE as a sigmoid activation enforces the output of
its decoder to belong to $[0,1]$.  We used three measures to conduct
this comparison:
\begin{itemize}
\item the reconstruction error, that is the pixel-wise mean squared
  error between the input images $\mathbf{x}^{(i)}$ of the test
  dataset and their reconstruction/approximation
  $\mathbf{\hat{x}}^{(i)}$:
  $\frac{1}{MN}\sum_{i=1}^M\sum_{j=1}^N(\mathbf{x}_j^{(i)} -
  \mathbf{\hat{x}}_j^{(i)})^2$;
\item the sparsity of the encoding, measured using the mean on all
  test images of the sparsity measure $\sigma$
  (Equation~\ref{eq:hoyer-sparsity}):
  $\frac{1}{M}\sum_{i=1}^M\sigma(\mathbf{h}^{(i)})$;
\item the approximation error to dilation by a disk of radius 1,
  obtained by computing the pixel-wise mean squared error between the
  dilation $\delta_B$ by a disk of radius 1 of the original image and
  the part-based approximation $D_B$ to the same dilation, using the
  learned representation:
  $\frac{1}{MN}\sum_{i=1}^M\sum_{j=1}^N (D_{B}(\mathbf{x}^{(i)})_j -
  \delta_B(\mathbf{x}^{(i)})_j)^2$.
\end{itemize}
 
\begin{figure}[hbpt]
  \captionsetup[subfloat]{farskip=0pt,captionskip=0pt}
  \centering
  \subfloat{\includegraphics[width=0.8\columnwidth]{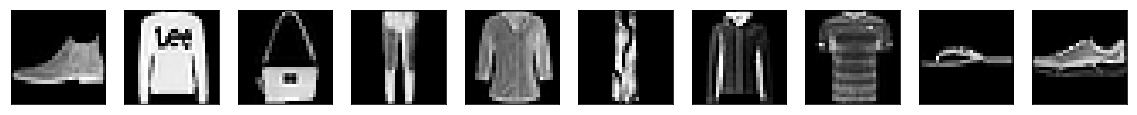}}
  
  \subfloat{\includegraphics[width=0.8\columnwidth]{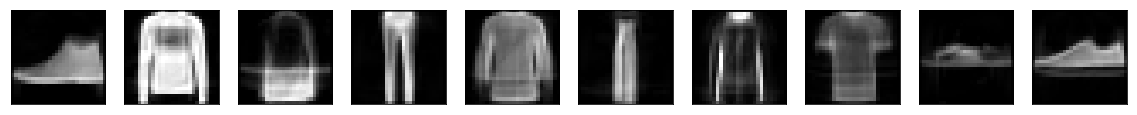}}
  
  \subfloat{\includegraphics[width=0.8\columnwidth]{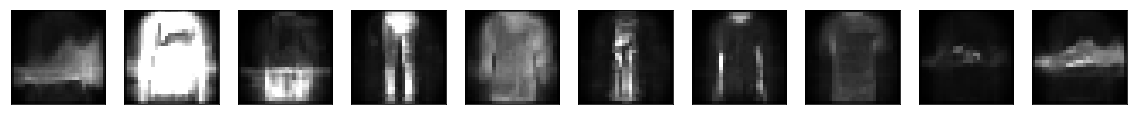}}
  
  \subfloat{\includegraphics[width=0.8\columnwidth]{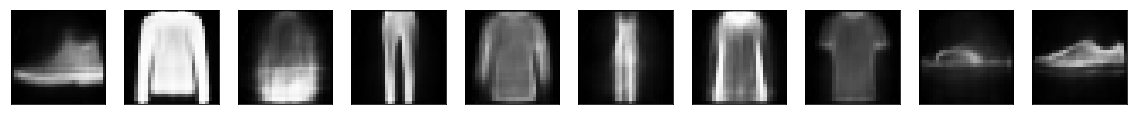}}
  
  \subfloat{\includegraphics[width=0.8\columnwidth]{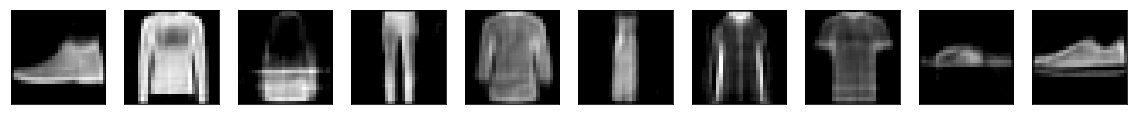}}
  \caption{Reconstruction of the Fashion-MNIST dataset (first row) by the sparse-NMF, the NNSAE, the NCAE and the AsymAE.}
  \label{fig:Fashion_MNIST_recs}
\end{figure}

The parameter settings used for NCAE and the NNSAE algorithms are the
ones provided in \cite{7310882,Lemme2012OnlineLA}. For the sparse-NMF,
a sparsity constraint of $S_h = 0.6$ was applied to the encodings and
no sparsity constraint was applied on the atoms of the
representation. For our AsymAE algorithm, $p=0.05$ was fixed for the
sparsity objective of the regularizer of
Equation~\eqref{eq:KL_div_penalty_function}, and the weight of the
sparsity regularizer in the loss function in
Equation~\eqref{eq:adding_sparse_regularization_term} was set to
$\beta=0.001$ for MNIST and $\beta=0.0005$ for Fashion-MNIST.  Various
other values have been tested for each algorithm, but the improvement
of one of the evaluation measures usually came at the expense of the
two others. Quantitative results are summarized in
Table~\ref{table:comparison_MNIST}. Reconstructions by the various
approaches of some sample images from both datasets are shown in
Figure~\ref{fig:Fashion_MNIST_recs}.

\begin{table}[htbp] 
  \caption{Comparison of the reconstruction error, sparsity of
    encoding and part-based approximation error to dilation produced
    by the sparse-NMF, the NNSAE, the NCAE and the AsymAE, for both
    MNIST and Fashion-MNIST datasets.}
\centerline{
\begin{tabular}{|c|c|c|c|}
  \hline
  Model & Reconstruction & Sparsity & Part-based approximation\\
        & error & of code & error to dilation\\ \hline
  \multicolumn{4}{c}{\textbf{MNIST}} \\ 
  \hline
  Sparse-NMF  & $0.011$ & $\mathbf{0.66}$ & $\mathbf{0.012}$\\
  \hline
  NNSAE & $0.015$ & $0.31$ & $0.028$\\  
  \hline
  NCAE & $0.010$ & $0.35$ & $0.18$\\
  \hline
  AsymAE & $\mathbf{0.007}$ & $0.54$ & $0.069$\\
  \hline
  \multicolumn{4}{c}{\textbf{Fashion MNIST}}\\ 
  \hline
  Sparse-NMF  & $0.011$ & $\mathbf{0.65}$ & $\mathbf{0.022}$\\
  \hline
  NNSAE & $0.029$ & $0.22$ & $0.058$\\  
  \hline
  NCAE & $0.017$ & $0.60$ & $0.030$\\
  \hline
  AsymAE & $\mathbf{0.010}$ & $0.52$ & $0.066$\\
  \hline
    \end{tabular}
    }
\label{table:comparison_MNIST}
\end{table}

\begin{figure}[htbp]
  \captionsetup[subfloat]{farskip=0pt,captionskip=0pt}
  \centering
  \subfloat[Sparse-NMF]{\includegraphics[width=0.23\columnwidth]{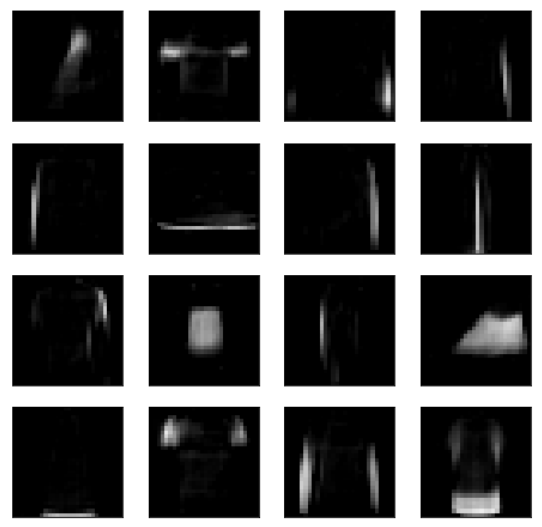}}
  \quad
  \subfloat[NNSAE]{\includegraphics[width=0.23\columnwidth]{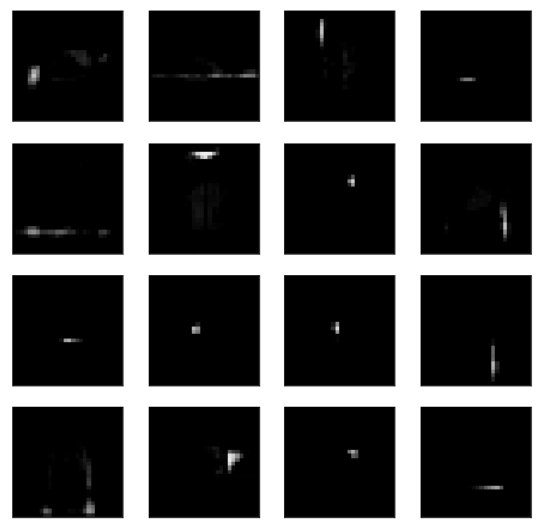}}
  \quad
  \subfloat[NCAE]{\includegraphics[width=0.23\columnwidth]{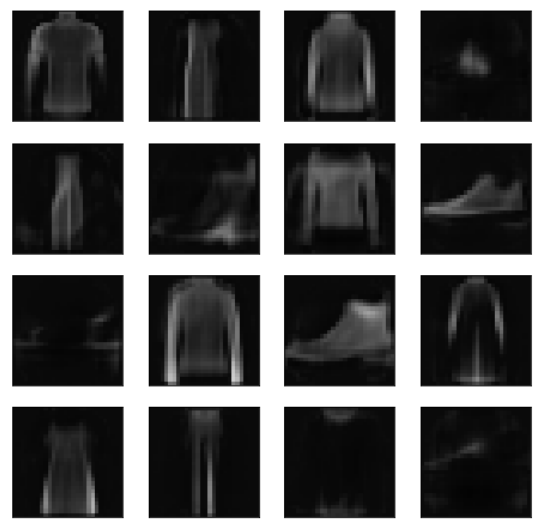}}
  \quad
  \subfloat[AsymAE]{\includegraphics[width=0.23\columnwidth]{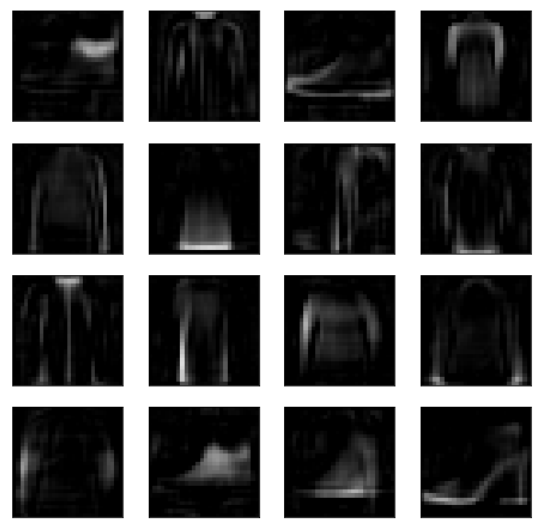}}
  \caption{16 of the 100 atom images of the four compared representations of Fashion-MNIST dataset.}
  \label{fig:Fashion_MNIST_atoms}
\end{figure}

\begin{figure}[htbp]
  \captionsetup[subfloat]{farskip=0pt,captionskip=0pt}
  \centering
  \subfloat{\includegraphics[width=0.8\columnwidth]{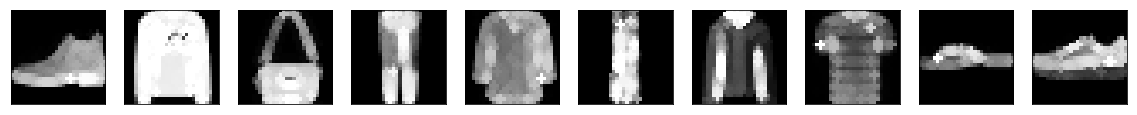}}
  
  \subfloat{\includegraphics[width=0.8\columnwidth]{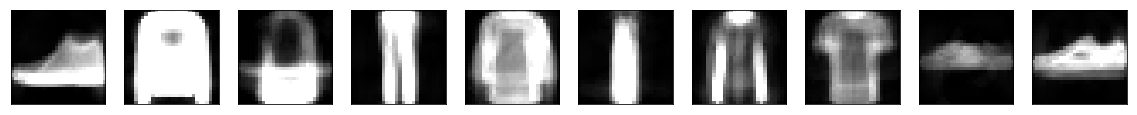}}
  
  \subfloat{\includegraphics[width=0.8\columnwidth]{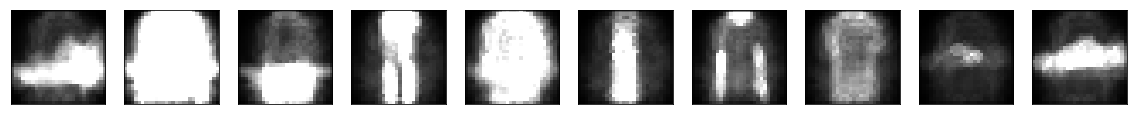}}
  
  \subfloat{\includegraphics[width=0.8\columnwidth]{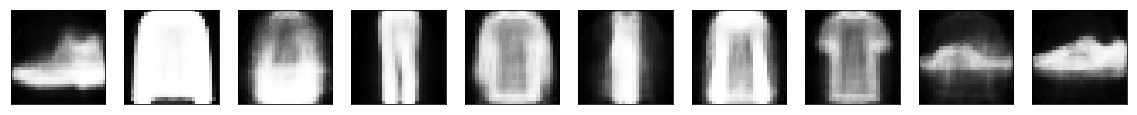}}
  
  \subfloat{\includegraphics[width=0.8\columnwidth]{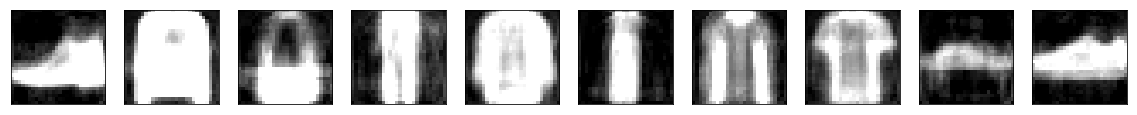}}
  \caption{Part-based approximation of the dilation by a structuring element of size 1 (first row), computed using the sparse-NMF, the NNSAE, the NCAE and the AsymAE.}
  \label{fig:Fashion_MNIST_max_approx}
\end{figure}

Both the quantitative results and the reconstruction images attest the
capacity of our model to reach a better trade-off between the accuracy
of the reconstruction and the sparsity of the encoding (that usually
comes at the expense of the former criteria), than the other neural
architectures. Indeed, in all conducted experiments, varying the
parameters of the NCAE and the NNSAE as an attempt to increase the
sparsity of the encoding came with a dramatic increase of the
reconstruction error of the model. We failed however to reach a
trade-off as good as the sparse-NMF algorithm that manages to match a
high sparsity of the encoding with a low reconstruction error,
especially on the Fashion-MNIST dataset.  The major difference between
the algorithms can be seen in Figure~\ref{fig:Fashion_MNIST_atoms}
that pictures 16 of the 100 atoms of each of the four learned
representations. While sparse-NMF manages, for both datasets, to build
highly interpretable and clean part-based representations, the two
deep baselines build representations that picture either too local
shapes, in the case of the NNSAE, or too global ones, in the case of
the NCAE. Our method suffers from quite the same issues as the NCAE,
as almost full shapes are recognizable in the atoms. We noticed
through experiments that increasing the sparsity of the encoding leads
to less and less local features in the atoms.  It has to be noted that
the $L_2$ Asymmetric Weight Decay regularization used by the NCAE and
NNSAE models allows for a certain proportion of negative weights. As
an example, up to $32.2\%$ of the pixels of the atoms of the NCAE
model trained on the Fashion-MNIST dataset are non-negative, although
their amplitude is lower than the average amplitude of the positive
weights. The amount of negative weights can be reduced by increasing
the corresponding regularization, which comes at the price of an
increased reconstruction error and less sparse encodings.  Finally
Figure~\ref{fig:Fashion_MNIST_max_approx} pictures the part-based
approximation to dilation by a structuring element of size one,
computed using the four different approaches on ten images from the
test set. Although the quantitative results state otherwise, we can
note that our approach yields a quite interesting part-based
approximation, thanks to a good balance between a low overlapping of
atoms (and dilated atoms) and a good reconstruction capability.

\vspace{-1cm}
\section{Conclusions and future works}
      
We have presented an online method to learn a part-based
dictionary representation of an image dataset, designed for
accurate and efficient approximations of morphological
operators. This method relies on auto-encoder networks, with a
deep encoder for a higher reconstruction capability and a
shallow linear decoder for a better interpretation of the
representation. Among the online part-based methods using
auto-encoders, it achieves the state-of-the-art trade-off
between the accuracy of reconstructions and the sparsity of
image encodings. Moreover, it ensures a strict (that is, non
approximated) non-negativity of the learned
representation. These results would need to be confirmed on
larger and more complex images (e.g. color images), as the
proposed model is scalable. We especially evaluated the learned
representation on an additional criterion, that is the
commutation of the representation with a morphological dilation,
and noted that all online methods perform worse than the offline
sparse-NMF algorithm. A possible improvement would be to impose
a major sparsity to the dictionary images an appropriate
regularization. Additionally, using a morphological
layer~\cite{Charisopoulos2017,ritter1996} as a decoder may be
more consistent with our definition of part-based approximation,
since a representation in the $(\max, +)$ algebra would commute
with the morphological dilation by essence.

\paragraph{Acknowledgments:} This work was partially funded by a grant
from Institut Mines-Telecom and MINES ParisTech.
%
%
%

\bibliographystyle{splncs04}
\bibliography{refCourt.bib}

\begin{thebibliography}{10}
\providecommand{\url}[1]{\texttt{#1}}
\providecommand{\urlprefix}{URL }
\providecommand{\doi}[1]{https://doi.org/#1}

\bibitem{angulo:hal-00658963}
Angulo, J., Velasco-Forero, S.: Sparse mathematical morphology using
  non-negative matrix factorization. In: Soille, P., Pesaresi, M., , Ouzounis,
  G.K. (eds.) 10th International Symposium on Mathematical Morphology and Its
  Application to Signal and Image Processing (ISMM). vol. LNCS 6671, pp. 1--12
  (2011)

\bibitem{DBLP:journals/corr/abs-1802-00003}
Ayinde, B.O., Zurada, J.M.: Deep learning of constrained autoencoders for
  enhanced understanding of data. CoRR  \textbf{abs/1802.00003} (2018)

\bibitem{Charisopoulos2017}
Charisopoulos, V., Maragos, P.: Morphological perceptrons: Geometry and
  training algorithms. pp. 3--15 (04 2017). \doi{10.1007/978-3-319-57240-6\_1}

\bibitem{infoGAN}
Chen, X., Duan, Y., Houthooft, R., Schulman, J., Sutskever, I., Abbeel, P.:
  Infogan: Interpretable representation learning by information maximizing
  generative adversarial nets. CoRR  \textbf{abs/1606.03657} (2016)

\bibitem{DBLP:journals/corr/DosovitskiySB14}
Dosovitskiy, A., Springenberg, J.T., Brox, T.: Learning to generate chairs with
  convolutional neural networks. CoRR  \textbf{abs/1411.5928} (2014)

\bibitem{7310882}
Hosseini-Asl, E., Zurada, J.M., Nasraoui, O.: Deep learning of part-based
  representation of data using sparse autoencoders with nonnegativity
  constraints. IEEE Transactions on Neural Networks and Learning Systems
  \textbf{27}(12),  2486--2498 (2016)

\bibitem{Hoyer2004}
Hoyer, P.O.: Non-negative matrix factorization with sparseness constraints.
  CoRR  \textbf{cs.LG/0408058} (2004)

\bibitem{batchnorm}
Ioffe, S., Szegedy, C.: Batch normalization: Accelerating deep network training
  by reducing internal covariate shift. CoRR  \textbf{abs/1502.03167} (2015)

\bibitem{lecun-mnisthandwrittendigit-2010}
LeCun, Y., Cortes, C.: {MNIST} handwritten digit database  (2010),
  \url{http://yann.lecun.com/exdb/mnist/}

\bibitem{NIPS2007_3313}
Lee, H., Ekanadham, C., Ng, A.Y.: Sparse deep belief net model for visual area
  v2. In: Platt, J.C., Koller, D., Singer, Y., Roweis, S.T. (eds.) Advances in
  Neural Information Processing Systems 20, pp. 873--880 (2008)

\bibitem{Lemme2012OnlineLA}
Lemme, A., Reinhart, R.F., Steil, J.J.: Online learning and generalization of
  parts-based image representations by non-negative sparse autoencoders. Neural
  Networks  \textbf{33},  194--203 (2012)

\bibitem{Maas2013RectifierNI}
Maas, A.L.: Rectifier nonlinearities improve neural network acoustic models.
  In: International Conference on Machine Learning (2013)

\bibitem{DBLP:journals/corr/MairalBP14}
Mairal, J., Bach, F.R., Ponce, J.: Sparse modeling for image and vision
  processing. CoRR  \textbf{abs/1411.3230} (2014)

\bibitem{maragos1986morphological}
Maragos, P., Schafer, R.: Morphological skeleton representation and coding of
  binary images. IEEE Transactions on Acoustics, Speech, and Signal Processing
  \textbf{34}(5),  1228--1244 (1986)

\bibitem{Morphological_profiles}
Pesaresi, M., Benediktsson, J.A.: A new approach for the morphological
  segmentation of high-resolution satellite imagery. IEEE Transactions on
  Geoscience and Remote Sensing  \textbf{39}(2),  309--320 (2001)

\bibitem{ritter1996}
Ritter, G., Sussner, P.: An introduction to morphological neural networks.
  vol.~4, pp. 709 -- 717 vol.4 (09 1996). \doi{10.1109/ICPR.1996.547657}

\bibitem{soille2013morphological}
Soille, P.: Morphological image analysis: principles and applications. Springer
  Science \& Business Media (2013)

\bibitem{Tanaka2003ColumnsFC}
Tanaka, K.: Columns for complex visual object features in the inferotemporal
  cortex: clustering of cells with similar but slightly different stimulus
  selectivities. Cerebral Cortex  \textbf{13 1},  90--9 (2003)

\bibitem{theis2005first}
Theis, F.J., Stadlthanner, K., Tanaka, T.: First results on uniqueness of
  sparse non-negative matrix factorization. In: 13th IEEE European Signal
  Processing Conference. pp.~1--4 (2005)

\bibitem{Angulo_Velasco_2017}
Velasco-Forero, S., Angulo, J.: Non-Negative Sparse Mathematical Morphology,
  vol.~202, chap.~1. Elsevier Inc.Academic Press (2017)

\bibitem{fashionMNIST}
Xiao, H., Rasul, K., Vollgraf, R.: Fashion-{MNIST}: a novel image dataset for
  benchmarking machine learning algorithms. arXiv:1708.07747  (2017)

\bibitem{7280364}
Zhang, L., Lu, Y.: Comparison of auto-encoders with different sparsity
  regularizers. In: International Joint Conference on Neural Networks (IJCNN).
  pp.~1--5 (2015)

\end{thebibliography}
\end{document}